\documentclass{article}
\PassOptionsToPackage{numbers, compress}{natbib}
\usepackage[preprint]{neurips_2023}
\usepackage[T1]{fontenc}

\usepackage[T1]{fontenc}    
\usepackage{hyperref}       
\usepackage{url}            
\usepackage{amsfonts}       
\usepackage{nicefrac}       
\usepackage{graphicx}
\usepackage{caption}
\usepackage{subcaption}
\usepackage{multirow}
\usepackage[ruled,noend,linesnumbered]{algorithm2e}
\usepackage{amssymb}
\usepackage{booktabs}
\usepackage[dvipsnames]{xcolor}
\usepackage{amsmath}

\usepackage{tikz}
\usetikzlibrary{quotes}
\usetikzlibrary{decorations.pathreplacing,calligraphy}\tikzset{operation/.style={draw, minimum size=0.5cm, circle,thick}}
\tikzset{func/.style={draw, minimum width=2cm, minimum height=1cm, rectangle,thick,align=center}}
\tikzset{alg/.style={draw, minimum width=2cm, minimum height=4cm, rectangle,dotted,align=center}}
\pgfmathsetmacro{\tikzHspace}{3}
\pgfmathsetmacro{\tikzVspace}{1.5}

\pgfmathsetmacro{\legendHspace}{.5}
\pgfmathsetmacro{\legendVspace}{-1.2}

\usepackage[textsize=tiny]{todonotes}

\newcommand{\mc}[1]{\mathcal{#1}}
\newcommand{\mi}[1]{\mathit{#1}}
\newcommand{\mf}[1]{\mathbf{#1}}

\newcommand{\jobs}{\mc{J}}
\newcommand{\jobI}[1]{\mi{J}_{#1}}
\newcommand{\nMachinesJobI}[1]{\mi{n}_{#1}}

\newcommand{\operations}{\mc{O}}
\newcommand{\opIJ}[2]{\mi{O}_{#1#2}}
\newcommand{\proctimeIJ}[2]{\mi{p}_{#1#2}}
\newcommand{\proctimeProbaIJ}[2]{\mathbb{P}_{#1#2}}

\newcommand{\machines}{\mc{M}}

\newcommand{\machineIJ}[2]{\mi{m}_{#1#2}}

\newcommand{\startdateIJ}[2]{\mi{S}_{#1#2}}
\newcommand{\completionIJ}[2]{\mi{C}_{#1#2}}
\newcommand{\makespan}{\mi{C}_{\mi{max}}}

\newcommand{\scheduledOps}{\mc{S}}
\newcommand{\actions}{\mc{A}}

\newcommand{\sol}{\sigma}

\newcommand{\graph}{\mc{G}}

\newcommand{\graphPrecArcs}{\mc{C}}
\newcommand{\graphMachinesEdges}{\mc{D}}

\newcommand{\graphMachinesEdgesDirected}{\graphMachinesEdges^O}

\newcommand{\reward}{\mi{r}}

\newcommand{\wheatNM}[2]{W-#1\textsf{x}#2}
\newcommand{\wheatdNM}[2]{Wd-#1\textsf{x}#2}
\newcommand{\ins}[2]{#1 \times #2}

\begin{document}	

\title{Learning to Solve Job Shop Scheduling under Uncertainty}

\author{%
  Guillaume Infantes\\
  Jolibrain, Toulouse, France\\
  \texttt{guillaume.infantes@jolibrain.com}
  \And
  Stéphanie Roussel\\
  ONERA-DTIS, Université de Toulouse, France\\
  \texttt{stephanie.roussel@onera.fr}\\
  \And
  Pierre Pereira\\
  Jolibrain, Toulouse, France\\
  \texttt{pierre.pereira@jolibrain.com}
  \And
  Antoine Jacquet\\
  Jolibrain, Toulouse, France\\
  \texttt{antoine.jacquet@jolibrain.com}
  \And
  Emmanuel Benazera\\
  Jolibrain, Toulouse, France\\
  \texttt{emmanuel.benazera@jolibrain.com}
}
\maketitle

\begin{abstract}
Job-Shop Scheduling Problem (JSSP) is a combinatorial optimization problem where tasks need to be scheduled on machines in order to minimize criteria such as makespan or delay. To address more realistic scenarios, we associate a probability distribution with the duration of each task. Our objective is to generate a robust schedule, i.e. that minimizes the average makespan. This paper introduces a new approach that leverages Deep Reinforcement Learning (DRL) techniques to search for robust solutions, emphasizing JSSPs with uncertain durations. Key contributions of this research include: (1) advancements in DRL applications to JSSPs, enhancing generalization and scalability, (2) a novel method for addressing JSSPs with uncertain durations. The Wheatley approach, which integrates Graph Neural Networks (GNNs) and DRL, is made publicly available for further research and applications.
\end{abstract}

\section{Introduction}

Job-shop scheduling problems (JSSPs) are combinatorial optimization problems that involve assigning tasks to resources (e.g., machines) in a way that minimizes criteria such as makespan, tardiness, or total flow time. While the scheduling of production resources plays an important role in many industries, the JSSPs formulation  lacks the handling of uncertainty  due to its  simplifying assumptions. This leads to several direct practical consequences, such as scheduling from fixed factors which can have significant impact on scheduling performance, and ignoring machine breakdowns or material shortages, leading to poor solutions when faced with real-world uncertainty.

Optimal solving of combinatorial optimization problems is NP-complete, and while there has been a lot of progress in solvers performance \cite{DACOL2022100249}, classical approaches remain often impractical on large instances. Therefore, approximation and heuristics-based methods have been proposed \cite{sels2012comparison}; but handling uncertainty within these methods remains challenging. Recent works have considered learning algorithms for such problems and report early advances with deep reinforcement learning (DRL) techniques (\cite{han-jssp-cnn,liu-jssp-cnn,zeng2022hybrid}). Because it models the world as a runnable environment, and the algorithm learns directly from it, DRL does offer a more natural way to handle uncertainty with JSSPs. As Reinforcement Learning methods are robust to noise, the uncertainty in the problem statement, which is reflected in the learner's environment, is naturally handled by the algorithm.

We present two contributions for tackling JSSP with uncertain durations.

First, this work shows a range of improvements over the DRL and JSSPs literature, from neural network architectures to training hyper-parameters and reward definitions. These directly lead to better generalization and scalability, both to same-size problems and  to larger problems. 

Second, the proposed method solves JSSPs with uncertain duration, that beats optimal deterministic solutions on expected uncertainty. This is relevant to the general use-case where uncertainty cannot be known in advance and where the best deterministic schedule uses expected uncertainty on tasks duration.

Overall, this leads to a very flexible and efficient approach, capable of naturally handling duration uncertainty, with top results on existing Taillard benchmarks while setting a new benchmark reference for JSSPs with uncertain durations.
The approach, code-named Wheatley, combines Graph Neural Networks (GNNs) and DRL techniques. The code is made available under an Open Source license at \url{https://github.com/jolibrain/wheatley/}.

The paper is organized as follows: related works are introduced in Section~\ref{sec:related_works}, then the JSSP with uncertainty is formalized as a Markov Decision Process (MDP) in Section~\ref{sec:JSSP_uncertainty}. In Section~\ref{sec:contrib}, we detail the core technical contributions. Section~\ref{sec:experiments} is dedicated to experiments on both deterministic and stochastic JSSPs. Finally, we conclude and discuss future works in Section~\ref{sec:conclusion}.

\section{Related work}

\label{sec:related_works}

This section provides an  overview of techniques developed to address both deterministic and stochastic versions of the Job hop Scheduling Problem (\cite{XIONG2022105731}).

\smallskip
\emph{Deterministic JSSPs.}
Mathematical programming, including techniques such as Constraint Programming (CP) or Integer Linear Programming (ILP), has been favored for solving JSSPs due to its precision and ability to model complex scheduling problems. However, the time and resources required to achieve  solutions can be very high for large scenarios \cite{DACOL2022100249}. 

Priority Dispatching Rules (PDRs), are heuristic-based strategies that assign priorities to jobs based on predefined criteria; they make local decisions at each step by picking the highest priority job. Common criteria used in PDRs include the Shortest Processing Time (jobs with the shortest processing time are given priority) and the Earliest Due Date (jobs due the earliest are prioritized). This simplifies the scheduling process, making PDRs particularly useful for real-time or large-scale scenarios where rapid decision-making is essential. However, while PDRs are computationally efficient, they rarely yield the optimal solution. A comprehensive evaluation  can be found in \cite{sels2012comparison}.

Recently, there has been a surge in machine learning and data-driven approaches to solve JSSPs. Instead of relying on handcrafted heuristics, the Learning to Dispatch strategy (L2D) uses machine learning to emulate successful dispatching strategies, as described in \cite{L2D}. A significant advantage of this method is its size-agnostic nature, as it uses the disjunctive graph representation of the JSSPs. Graph Neural Networks (GNNs) process these graphs to capture intricate relations between operations and their constraints. Deep Reinforcement Learning (DRL) guides the decision-making process, optimizing scheduling decisions based on the features extracted by the GNNs.
In \cite{park-gnn-rl-jssp}, the authors leverage a GNN to convert the JSSP graph into node representations. These representations  assist in determining scheduling actions. Proximal Policy Optimization (PPO) is employed as a training method for the GNN-derived node embeddings and the associated policy. This method employs an event-based simulator for the JSSP and directly incorporates times into the states of the base Markov Decision Process. This specificity complicates its adaptation to uncertain scenarios. \cite{flexible-jssp} also uses GNNs and PPO for addressing the flexible job-shop problem where the agent also has to choose machines for tasks; the authors add nodes for machines and use two different types of message-passing. The same problem is addressed using a bipartite graph and custom message passing  in \cite{rev7}, with good results. 
The Reinforced Adaptive Staircase Curriculum Learning (RASCL) approach \cite{IklassovMR023} is a  Curriculum Learning method that adjusts difficulty levels during learning by dynamically focusing on  challenging instances. 

\smallskip
\emph{Robustness in JSSPs.}
Stochastic JSSPs (SJSSP) account for uncertainties in processing times  by modeling them as random variables. 
Some techniques  use classic solvers to generate robust solutions by anticipating potential disruptions or modeling worst-case scenarios. For instance, in \cite{LocalSolverStocJSSP}, for a given JSSP, several processing times scenarios  are sampled. The objective is therefore to generate a unique schedule that is good for all sampled scenarios. PDRs can also be used but they can lack a global view, as in the deterministic case. 
Several works address SJSSP through meta-heuristics, as described in \cite{bianchi2009survey}. Other techniques involving genetic algorithms and their hybridization are also widely used \cite{boukedroun2023hybrid,rev2}. \cite{rev5} introduces a way to robustify solution to deterministic relaxation of the SJSSP. \cite{rev6} presents a both proactive and reactive scheduling: a multi-agent architecture is responsible for the proactive robust scheduling and a repair procedure is involved for machine breakdown and arrival of rush jobs. 

Some works address the dynamic variant of SJSSP, in which new jobs can arrive at any time. \cite{rev3} is a review of such extensions along with corresponding proposed solving methods. Classical approaches involve complex mathematical programming models that do not scale well \cite{rev1}. Online reactive recovery approaches are a possible solution  if no robust solution is available \cite{rev4}.   More recent approaches explore the use of using DRL and GNN \cite{liu2023dynamic}. However, to the best of our knowledge, there are no work that use such techniques for the SJSSP.

\section{JSSP with Uncertainty as a MDP}
\label{sec:JSSP_uncertainty}

In this section, we first recall the JSSP definition. Then, we describe how to represent uncertainty and define the corresponding MDP. We finally present how to use Reinforcement Learning (RL) for solving the MDP.

\subsection{Background}
\label{sec:background}

A JSSP is defined as a pair $(\jobs,\machines)$, where $\jobs$ is a set of jobs and $\machines$
is a set of machines. Each job $\jobI{i} \in \jobs$ must go through $\nMachinesJobI{i}$ machines in $\machines$ in a given order  $\opIJ{i}{j} \to ... \to \opIJ{i}{\nMachinesJobI{i}}$, where each element $\opIJ{i}{j} (1 \leq j \leq \nMachinesJobI{i} )$ is called an operation of $\jobI{i}$. The binary relation $\to$ is a  precedence constraint. 
The size of a JSSP instance is classically denoted as $|\jobs|\times|\machines|$. 
 In the following, the set of all operations is denoted $\operations$. To be executed, each operation $\opIJ{i}{j} \in \operations$ requires a unique machine $\machineIJ{i}{j} \in \machines$ during a processing time denoted $\proctimeIJ{i}{j}$ ($\proctimeIJ{i}{j} \in \mathbb{N}^+$). Each machine can only process one job at a time, and preemption is not allowed.
 
A \emph{solution} $\sol$ of a JSSP instance is a function that assigns a start date $\startdateIJ{i}{j}$ to each operation $\opIJ{i}{j}$  so that precedence between operations of each job are respected and there is no temporal overlap between operations that are performed on the same machine. The completion time of an operation $\opIJ{i}{j}$, is  $\completionIJ{i}{j} = \startdateIJ{i}{j} + \proctimeIJ{i}{j}$. A solution $\sigma$ is \emph{optimal} if it minimizes the makespan  $\makespan = \mi{max}_{\opIJ{i}{j} \in \operations} \{\completionIJ{i}{j}\}$, \textit{i.e.} the maximal completion time of operations. 
 
 As described in \cite{roy1964problemes}, the disjunctive graph is defined by $\graph = (\operations, \graphPrecArcs, \graphMachinesEdges)$ of a JSSP $(\jobs,\machines)$  as:
	\begin{itemize}
			\item $\operations$ is the set of vertices, i.e. there is one vertex for each operation $o \in \operations$; 
			\item $\graphPrecArcs$ is a set of directed arcs representing the precedence constraints between operations of each job (conjunctions);
			\item $\graphMachinesEdges$ is a set of edges (disjunctions), each of which connects a pair of operations requiring the same
			machine for processing.
	\end{itemize}
	
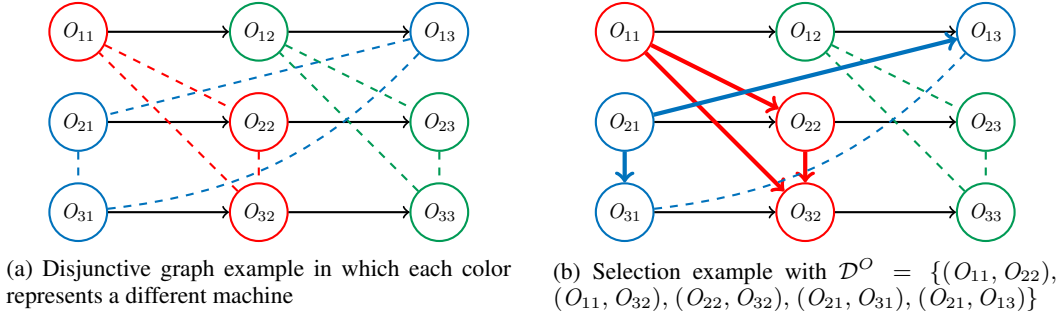
\begin{figure}[t]
	\centering
	\begin{subfigure}[t]{0.48\textwidth}
		\centering
	\begin{tikzpicture}[scale=0.8, every node/.style={transform shape}]
		\node[operation,draw=red] (o11) at (0,0)  {$\opIJ{1}{1}$};
		\node[operation,draw=ForestGreen] (o12) at (\tikzHspace,0) {$\opIJ{1}{2}$};
		\node[operation,draw=RoyalBlue] (o13) at (2*\tikzHspace,0) {$\opIJ{1}{3}$};
		
		\node[operation,draw=RoyalBlue] (o21) at (0,-\tikzVspace) {$\opIJ{2}{1}$};
		\node[operation,draw=red] (o22) at (\tikzHspace,-\tikzVspace) {$\opIJ{2}{2}$};
		\node[operation,draw=ForestGreen] (o23)  at (2*\tikzHspace,-\tikzVspace) {$\opIJ{2}{3}$};
		
		\node[operation,draw=RoyalBlue] (o31) at (0,-2*\tikzVspace) {$\opIJ{3}{1}$};
		\node[operation,draw=red] (o32) at (\tikzHspace,-2*\tikzVspace) {$\opIJ{3}{2}$};
		\node[operation,draw=ForestGreen] (o33)  at (2*\tikzHspace,-2*\tikzVspace) {$\opIJ{3}{3}$};

		\foreach \i in {1,2,3} {
			\draw[->,thick] (o\i1) -- (o\i2);
			\draw[->,thick] (o\i2) -- (o\i3);
		}
		
		\draw[thick,red,dashed] (o11) -- (o22) -- (o32) -- (o11);
		
		\draw[thick,ForestGreen,dashed] (o12) -- (o23) -- (o33) -- (o12);
		\draw[thick,RoyalBlue,dashed] (o13) -- (o21) -- (o31);
		\path[thick,RoyalBlue,dashed] (o31) edge [bend right=20] (o13);
	\end{tikzpicture}
	\caption{Disjunctive graph example in which each color represents a different machine}
	\label{fig:disj}
	\end{subfigure}
	\hfill
	\begin{subfigure}[t]{.48\textwidth}
		\centering
		\begin{tikzpicture}[scale=0.8, every node/.style={transform shape}]
			\node[operation,draw=red] (o11) at (0,0)  {$\opIJ{1}{1}$};
			\node[operation,draw=ForestGreen] (o12) at (\tikzHspace,0) {$\opIJ{1}{2}$};
			\node[operation,draw=RoyalBlue] (o13) at (2*\tikzHspace,0) {$\opIJ{1}{3}$};
			
			\node[operation,draw=RoyalBlue] (o21) at (0,-\tikzVspace) {$\opIJ{2}{1}$};
			\node[operation,draw=red] (o22) at (\tikzHspace,-\tikzVspace) {$\opIJ{2}{2}$};
			\node[operation,draw=ForestGreen] (o23)  at (2*\tikzHspace,-\tikzVspace) {$\opIJ{2}{3}$};
			
			\node[operation,draw=RoyalBlue] (o31) at (0,-2*\tikzVspace) {$\opIJ{3}{1}$};
			\node[operation,draw=red] (o32) at (\tikzHspace,-2*\tikzVspace) {$\opIJ{3}{2}$};
			\node[operation,draw=ForestGreen] (o33)  at (2*\tikzHspace,-2*\tikzVspace) {$\opIJ{3}{3}$};

			\foreach \i in {1,2,3} {
				\draw[->,thick] (o\i1) -- (o\i2);
				\draw[->,thick] (o\i2) -- (o\i3);
			}

			\draw[ultra thick,red,->] (o11) -- (o22);
			\draw[ultra thick,red,->] (o11) -- (o32);
			\draw[ultra thick,red,->] (o22) -- (o32);
			
			\draw[thick,ForestGreen,dashed] (o12) -- (o23) -- (o33) -- (o12);
			
			\draw[ultra thick,RoyalBlue,->] (o21) -- (o31);
			\draw[ultra thick,RoyalBlue,->] (o21) -- (o13);
			
			\draw[thick,RoyalBlue,dashed] (o13) -- (o21) -- (o31);
			\path[thick,RoyalBlue,dashed] (o31) edge [bend right=20] (o13);
		\end{tikzpicture}
		\caption{Selection example with $\graphMachinesEdgesDirected = \{(\opIJ{1}{1}, \opIJ{2}{2}),$ $(\opIJ{1}{1}, \opIJ{3}{2}),$ $(\opIJ{2}{2}, \opIJ{3}{2}),$ $(\opIJ{2}{1}, \opIJ{3}{1}),$ $(\opIJ{2}{1}, \opIJ{1}{3})\}$ }
		\label{fig:disj_with_selection}
	\end{subfigure}
	\caption{Disjunctive graph representation}
\end{figure}

Figure~\ref{fig:disj} shows the disjunctive graph of a JSSP with 3 jobs and 3 machines.
A \emph{selection} is a state of the graph in which a direction is chosen for some edges in $\graphMachinesEdges$, denoted $\graphMachinesEdgesDirected$. If an edge $(\opIJ{i}{j}, \opIJ{i'}{j'})$ in $\graphMachinesEdges$ becomes oriented (in that order), then it represents that operation $\opIJ{i}{j}$ is performed before $\opIJ{i'}{j'}$ on their associated machine. 
A selection is valid if the set of oriented arcs ($\graphPrecArcs \cup \graphMachinesEdgesDirected$) makes the graph acyclic. 
A solution $\sigma$ can be defined by a valid selection in which all edges in $\graphMachinesEdges$ have a direction ($\graphMachinesEdges = \graphMachinesEdgesDirected$) and in which start dates of operations are the earliest possible dates consistent with the selection precedences, as done in a classical Schedule Generation Scheme (SGS). 
Figure~\ref{fig:disj_with_selection} illustrates a valid selection for the toy JSSP instance of Figure~\ref{fig:disj}.

\subsection{Representing uncertainty}
JSSPs can be easily extended with duration uncertainty as bounds on tasks' duration, and effect uncertainty as failure outcomes of a task. While task failures could be represented using special nodes representing completely different outcomes, this would push the boundaries outside JSSP formal capabilities. In this work, failures are handled as retries that consume an uncertain time duration a fixed maximum number of times. In the following, we focus on uncertain task duration as a generic enough scheme to capture relevant uncertainty use-cases.

In the classical JSSP definition, every operation $\opIJ{i}{j}$ has a deterministic processing time $\proctimeIJ{i}{j}$. We extend this definition by saying that processing times are not known in advance, but that each operation $\opIJ{i}{j}$ has an associated probability distribution $\proctimeProbaIJ{i}{j}$ over its possible duration values.
The objective is to minimize the average makespan, that is formally defined by $\mi{max}_{\opIJ{i}{j} \in \operations} \int_{0}^{+\infty} (\startdateIJ{i}{j} + \mi{p_{ij}}) \, d\proctimeProbaIJ{i}{j}$.

\subsection{Sequential Decision Making}

The scheduling problem boils down to a Markov Decision Process.
Inputs of the problem are the the original disjunctive graph $\graph = (\operations, \graphPrecArcs, \graphMachinesEdges)$ and the probability distribution over duration values $\proctimeProbaIJ{i}{j}$ associated with each operation $\opIJ{i}{j}$.

\paragraph{State} The state $s_t$ at decision step $t$ is defined by:
\begin{itemize}
	\item the current selection $\graphMachinesEdgesDirected_t$,
	\item the set of already scheduled operations $\scheduledOps_t$ at step $t$.
\end{itemize}

The initial state $s_0$ is the disjunctive graph representing the original JSSP instance with $\graphMachinesEdgesDirected_0 = \emptyset$ and $\scheduledOps_0 = \emptyset$. The terminal state $s_T$ is a complete solution where $\graphMachinesEdgesDirected_T = \graphMachinesEdges$, \textit{i.e.} all disjunctive arcs have been assigned a direction, and $\scheduledOps_T = \operations$, \textit{i.e.} all operations have been scheduled.

\paragraph{Actions}

At each step $t$, candidate actions consist in selecting an operation $\opIJ{i}{j}$ to put directly after the last scheduled operation on the corresponding machine. This is a simple way to ensure that cycles are never added in $\graph$. Intuitively, it consists in choosing an operation to do before all the ones that have not yet been scheduled, and update the current selection accordingly. Furthermore, we force that an operation is a candidate for selection only if its preceding tasks in the same job have been scheduled. As exactly one operation is scheduled at each step, the final state is reached at step $T = \mi{card}(\operations)$.

Candidates actions $\actions_t$ at step $t$ are formally defined by $\actions_t = \{\opIJ{i}{j} \in \operations \;|\; \opIJ{i}{j} \notin \scheduledOps_t \textit{ and } \forall j' < j, \opIJ{i}{j'} \in \scheduledOps_t\}$.

\paragraph{Transitions} 
 If the chosen action at step $t$ consists in selecting the operation $\opIJ{i}{j}$ in $\actions_t$, then it  leads to adding $\opIJ{i}{j}$ to the scheduled operations set and adding the arc $(\opIJ{k}{l},\opIJ{i}{j})$ to $\graphMachinesEdgesDirected$ for each operation $\opIJ{k}{l}$ scheduled at step $t$ on the same machine. Formally, this gives:
 \begin{itemize}
 	\item $\scheduledOps_{t+1} = \scheduledOps_{t} \cup \{\opIJ{i}{j}\}$
 	\item $\graphMachinesEdgesDirected_{t+1} = \graphMachinesEdgesDirected_t \cup \{(\opIJ{k}{l}, \opIJ{i}{j}) \in \graphMachinesEdges \; | \; \opIJ{k}{l} \in \scheduledOps_{t} \textit{ and } \machineIJ{k}{l} = \machineIJ{i}{j}\}$
 \end{itemize}

\paragraph{Reward/Cost}
In most approaches to solve MDPs with reinforcement learning, it is preferred to use non-sparse rewards, \textit{i.e.} an informative reward signal at  every step. For instance  the authors of \cite{L2D} propose to use the difference of makespan induced by the affectation of the task, as this naturally sums to makespan at the end of the trajectory. With the presence of uncertainty, while we could compute bounds on the makespan in the same way, it is not obvious to aggregate such bounds into a uni-dimensional reward signal.

We use a different approach: we draw durations only when the schedule is complete (at time $T$) and give it as a cost. Start date of each operation is its earliest possible date considering conjunctive and disjunctive precedence arcs in the solution. All other rewards are null. Formally, it is defined as follows:
\begin{itemize}
	\item $\forall t < T$, $\reward_t = 0$;
	\item $\reward_T = \mi{max}_{\opIJ{i}{j} \in \operations} (\startdateIJ{i}{j} + \proctimeIJ{i}{j}^\mi{sample})$ where $\proctimeIJ{i}{j}^\mi{sample} \sim \proctimeProbaIJ{i}{j}$. 
\end{itemize}

As reinforcement learning  aims at minimizing the expectation of costs sum along trajectories, this corresponds to our objective of minimizing the average makespan.

\subsection{Solving the MDP with Reinforcement Learning}
We use a reinforcement learning setup, where the agent selects tasks to schedule, and gets corresponding partial schedules as observation along with rewards. Using this modeling, effects of actions are deterministic (as they only add edges in the graph), all uncertainty is in the reward value.  

The objective is to find a policy that minimizes the average makespan over a set of test problems that are not used during the training phase. To do so, we use a simulator that generates problems that are close to the test problems, and aim at obtaining a policy that minimizes the expectation of makespan along the problems generated by the simulator. Such a policy has to be able to generalize to test problems, \textit{i.e.} give good results without further learning. As the only source of uncertainty is the durations for which parameters only are observable, we want our parametric policy to be able to adapt to these parameters.

\paragraph{Algorithm}

In order to learn a policy, we consider a parametric policy and use the  Proximal Policy Optimization (PPO) algorithm \cite{PPO}, with action masking \cite{ppo_mask}. PPO is an on-policy actor-critic RL algorithm. Its current stochastic policy is the actor, while the critic estimates the quality of the current state.
More precisely, as shown in Algorithm~\ref{algo:general_algo}, the algorithm starts by randomly initializing parameters of policy (actor) and value function estimator (critic). Then, for a given number of iterations, its starts by collecting trajectory data in the form (observation, action, next observation) on train problem instances where actions are chosen using current stochastic policy. It then computes makespans  corresponding to train instances and by sampling durations and applying chosen order of actions. Using this, it computes returns at every timestep, then advantages (difference of sampled returns to value estimation) using current critic. Observed graphs are then rewired  (see section \ref{sec:graph-embedding}).
The PPO update algorithm itself samples a subset of corresponding observations, actions, advantages, value prediction and returns, then updates the actor parameters (including GNN) using the gradient of the advantages, and updates the critic (including GNN) using gradient of mean square error between the critic value and the observed returns. This PPO update is repeated a small number of times or until the variation in the policy would lead to out-of-distribution critic estimations (because samples are collected using the old policy, i.e. the one before PPO updates).

We then evaluate the current policy (actor) by playing the argmax of the stochastic policy on a given set of validation problems. We repeat these steps until the policy does not seem to improve on validation instances (the $N$ in the external for loop is an upper bound on the number of iterations, which is set to a large value and the for loop is interrupted manually).

	\begin{algorithm}
         \caption{General algorithm\label{algo:general_algo}}
		\DontPrintSemicolon
		Generate validation instances, compute heuristic and ortools performance on these instances \;
		\tcp{actor is $\thicksim$ current policy $\pi_{\theta}$}
		Init actor \;
		\tcp{critic is $\thicksim$ value function estimator}
		Init critic \;
		\For{$i = 1, 2, \ldots N$}{
			\tcp{Collect dataset}
			Generate train instances  \;
			Collect trials data $\mathcal{D}_i = ((s_t, a_t, r_t, s_{t+1}), ...)$ using current actor \;
			For each  trial : sample makespan using system simulator ($=$ final cost) \;
			Compute returns on trials \;
			Compute advantages on trials using current critic \;
			Rewire graphs in trial data \;
			\tcp{PPO update algorithm}
			\Repeat{max number of iterations or too large KL-divergence between current and updated policy}{
				Sample a minibatch of n data points over shuffled collected data \;
				Update actor over the minibatch data towards advantage maximization \;
				Update critic by MSE regression \;
			}
			Evaluate current policy (actor) on validation instances \;
		}
	\end{algorithm}

\section{GNN Implementation: Rewiring, Embedding and Addressing Uncertainty}
\label{sec:contrib}

\begin{figure}[!htbp]
  \centering
  \begin{tikzpicture}[scale=.9,every text node part/.style={align=center},every node/.style={scale=0.9}]
  	\tikzstyle{elem} = [draw=none,fill=none,text width=1.8cm]
  	
  	\begin{scope}
  		\node[draw,rounded corners, rectangle, minimum width=10cm, minimum height=3cm,fill=black!5]  (agent) at (4,0) {};
  		
  		\node[elem] (graphRewirer) at (0,0) {Graph\\Rewirer};
  		\node[elem] (nodeEmbed) at (3,.75) {Node\\Embedder};
  		\node[elem] (edgeEmbed) at (3,-.75) {Edge\\Embedder};
  		\node[elem,text width=1.2cm] (gnn) at (5.5,0) {GNN};
  		\node[elem] (valEstim) at (8,.75) {Value\\Estimator};
  		\node[elem] (actSelector) at (8,-.75) {Action\\Selector};
  		
  		\node at (0,-.95) {\includegraphics[width=1.3cm]{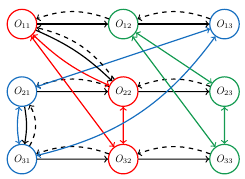}};
  		
  		\draw[->] (graphRewirer) -- (nodeEmbed);
  		\draw[->] (graphRewirer) -- (edgeEmbed);
  		\draw[->] (nodeEmbed) -- (gnn);
  		\draw[->] (edgeEmbed) -- (gnn);
  		\draw[->] (gnn) -- (valEstim) node [pos=.3, above] {\footnotesize graph\\ \footnotesize logit};
  		\draw[->] (gnn) -- (actSelector) node [pos=.3, below] {\footnotesize nodes\\ \footnotesize logits};;

  		\node at (-0.4,1.1) {Agent};
  		\draw[dashed] (-1,0.75) -- (.2,0.75);
  		\draw[dashed] (.2,0.75) -- (.2,1.5);
  	\end{scope}
  	
  	\begin{scope}[yshift=-5cm,xshift=2.5cm]
  		\node[draw,rounded corners, rectangle, minimum width=6cm, minimum height=3cm, fill=black!5] (simu)  at (1.5,0) {};

  		\node[elem] (graphUpdate) at (0,0) {Graph Update};
  		\node[elem,text width=2.2cm] (sysSimu) at (3,0) {System Simulator (uncertainty)};
  		
  		\begin{scope}[xshift=-.1cm]
  			\draw[red] (-.3,-1) circle (3pt);
  			\draw[red] (-.7,-.6) circle (3pt);
  			\draw[ForestGreen] (-.3,-.6) circle (3pt);
  			\draw[RoyalBlue] (-.7,-1) circle (3pt);
  			\draw (-.61,-1) -- (-.39,-1);
			\draw (-.61,-.6) -- (-.39,-.6);
  			\draw[dashed,red] (-.63,-.67) -- (-.37,-.93);
  		\end{scope}
  	
  		\node at (0,-.8) {\footnotesize $\Rightarrow$};

		\begin{scope}[xshift=1.1cm]
			\draw[red] (-.3,-1) circle (3pt);
			\draw[red] (-.7,-.6) circle (3pt);
			\draw[thick,->,red] (-.63,-.67) -- (-.37,-.93);
			\draw[ForestGreen] (-.3,-.6) circle (3pt);
			\draw[RoyalBlue] (-.7,-1) circle (3pt);
			\draw (-.61,-1) -- (-.39,-1);
			\draw (-.61,-.6) -- (-.39,-.6);
		\end{scope}

  		\node at (-0.6,1.1) {Simulator};
  		\draw[dashed] (-1.5,0.75) -- (.3,0.75);
  		\draw[dashed] (.3,0.75) -- (.3,1.5);
  	\end{scope}
  	
  	\draw (8.6,-.75) -- (9.5,-.75) -- (9.5,-5) node [midway, right] {Action\\$O_{ij}$};
  	\draw[->] (9.5,-5) -- (simu.east);
  	
  	\draw (simu.west) -- (-1.5,-5) -- (-1.5,0) node [pos=.7, left] {Partial\\Schedule\\Graph};
  	\draw[->] (-1.5,0) -- (agent.west);
  	
  	\node[anchor=east] at (-1.5,-3) {\includegraphics[width=1.5cm]{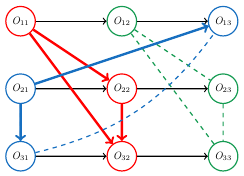}};
  	
  	\node[text centered, rounded corners, dotted, rectangle, draw, text width=1cm, minimum height=.75cm, fill=black!5] (ppo) at (4,-2.5)  {PPO};
  	
  	\draw[->,dotted] (9.5,-2.5) -- (ppo.east);
  	\draw[->,dotted] (-1.5,-2.5) -- (ppo.west);
  	
  	\draw[->,dotted] (ppo) -- (edgeEmbed);
  	\draw[->,dotted] (ppo) -- (nodeEmbed) node[pos=.15] {\footnotesize update parameters};
  	\draw[->,dotted] (ppo) -- (gnn);
  	\draw[->,dotted] (sysSimu) -- (ppo) node[pos=.8,right] {\footnotesize reward};
  	
  	\draw[dotted] (8.6,.75) -- (10,.75) -- (10,-2.3);
  	\draw[->,dotted] (10,-2.3) -- (4.625, -2.3) node[pos=.8,above] {\footnotesize value};
  	
  \end{tikzpicture}
  \caption{General Architecture}
  \label{fig:archi}
\end{figure}
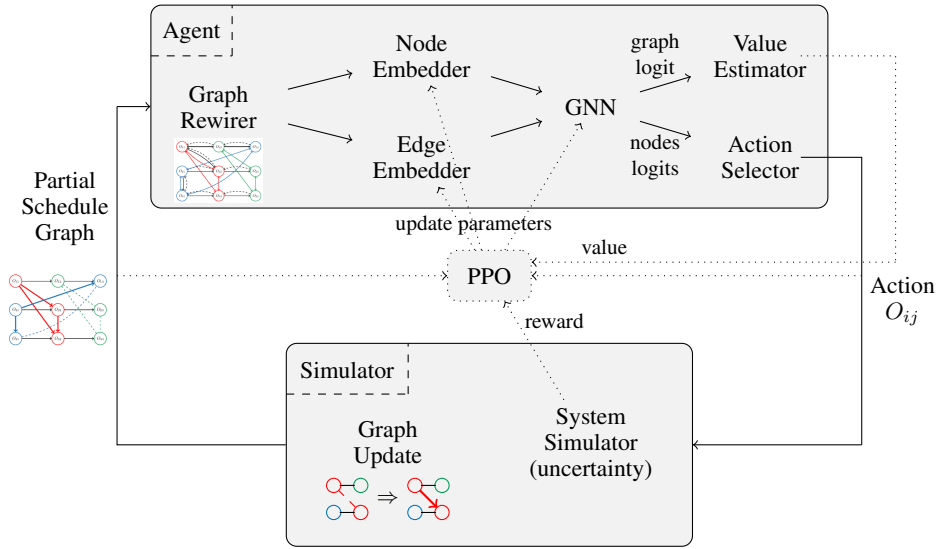

An overview of the architecture is shown in Figure~\ref{fig:archi}. The agent takes as an input a partial schedule in the form of a graph, as in Figure~\ref{fig:disj_with_selection}. Several elements, described in this section, are within the actor and allow to choose one action. This action is treated by the simulator to update the schedule graph by adding arcs and simulates the uncertainty when the last state is reached. Note that PPO uses the schedule graph, the action, the reward and the value estimation in order to update the embedders and the GNN. 

\subsection{Graph Rewiring and Embedding}
\label{sec:graph-embedding}
The representation above allows to model partial schedules (where some conflicts are not resolved) as disjunctive graph representation. 
Generally speaking, Message-Passing Graph Neural Networks (MP-GNN) use the graph structure as a computational lattice, meaning that information has to follow the graph adjacencies and only them. We thus have to make the difference between the input graph and the graph used by the MP-GNN. This is known as ``graph rewiring'' in the MP-GNN literature. In our case, if we use only precedencies as adjacencies, this would mean the we explicitly forbid information to go from future tasks to present choice of dispatch, which is definitely not what we want: we want the agent to choose task to dispatch based on effects on future conflicts, meaning that we want information go from future to present task.

\paragraph{Precedences.}
In order to have a rewired graph as small as possible, we remove from $\graphMachinesEdgesDirected$ all edges that are not necessary to obtain the complete order. For instance, we remove  from figure \ref{fig:disj_with_selection}, the edge $(\opIJ{1}{1}, \opIJ{3}{2})$. 
Links are then  added in the rewired  graph in both directions for every precedence, with different types for precedence and reverse-precedence edges. This enables learned operators to differentiate between chronological and reverse-chronological links and allows the network to pass information in a forward and backward way, depending on what is found useful during learning phase.

\paragraph{Conflicts.}

Remains the challenge of allowing message circulation between tasks sharing the same machine in the GNN. 
Two options are possible: 1) adding a node representing a machine with links to tasks using the machine and edges in both directions (from tasks to this machine node and in the opposite direction), or 2) directly connecting tasks that share a machine, resulting in a clique per machine in the message-passing graph.  In this paper, we choose the second approach as it showed better results than the first one in preliminary experiments. 

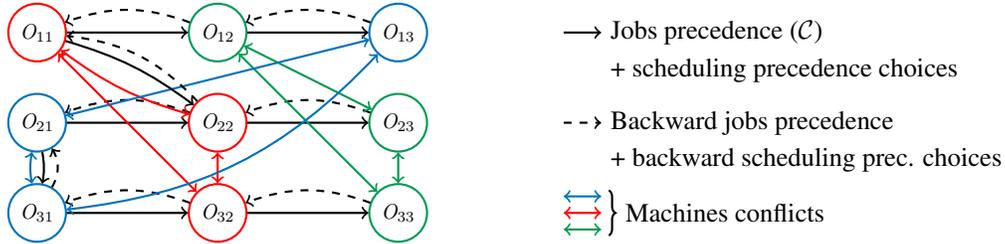
\begin{figure}[!htbp]
  \begin{tikzpicture}
  	\begin{scope}[scale=.8, transform shape]
  		\node[operation,draw=red] (o11) at (0,0)  {$\opIJ{1}{1}$};
  		\node[operation,draw=ForestGreen] (o12) at (\tikzHspace,0) {$\opIJ{1}{2}$};
  		\node[operation,draw=RoyalBlue] (o13) at (2*\tikzHspace,0) {$\opIJ{1}{3}$};
  		
  		\node[operation,draw=RoyalBlue] (o21) at (0,-\tikzVspace) {$\opIJ{2}{1}$};
  		\node[operation,draw=red] (o22) at (\tikzHspace,-\tikzVspace) {$\opIJ{2}{2}$};
  		\node[operation,draw=ForestGreen] (o23)  at (2*\tikzHspace,-\tikzVspace) {$\opIJ{2}{3}$};
  		
  		\node[operation,draw=RoyalBlue] (o31) at (0,-2*\tikzVspace) {$\opIJ{3}{1}$};
  		\node[operation,draw=red] (o32) at (\tikzHspace,-2*\tikzVspace) {$\opIJ{3}{2}$};
  		\node[operation,draw=ForestGreen] (o33)  at (2*\tikzHspace,-2*\tikzVspace) {$\opIJ{3}{3}$};
  		\foreach \i in {1,2,3} {
  			\draw[->,thick] (o\i1) -- (o\i2);
  			\draw[->,dashed,thick] (o\i2) edge [bend right=20] (o\i1);                
  			\draw[->,thick] (o\i2) -- (o\i3);
  			\draw[->,dashed, thick] (o\i3) edge [bend right=20] (o\i2);
  		}
  		
  		\draw[<->, thick, red] (o11) edge [bend right=10] (o22);
  		\draw[<->, thick, red] (o22) edge (o32);
  		\draw[<->, thick, red] (o32) edge  (o11);
  		
  		\draw[<->, thick, ForestGreen] (o12) edge (o23);
  		\draw[<->, thick, ForestGreen] (o23) edge (o33);
  		\draw[<->, thick, ForestGreen] (o33) edge (o12);
  		
  		\draw[<->, thick, RoyalBlue] (o13) edge (o21);
  		\draw[<->, thick, RoyalBlue] (o21) edge [bend right=10] (o31);
  		\draw[<->, thick, RoyalBlue] (o31) edge [bend right=20] (o13);
  		
  		\draw[->, thick] (o11) edge [bend left=10] (o22);
  		\draw[->, thick, dashed] (o22) edge [bend right=20] (o11);
  		\draw[->, thick] (o21) edge [bend left=10] (o31);
  		\draw[->, thick, dashed] (o31) edge [bend right=30] (o21);
  	\end{scope}

			  \begin{scope}[xshift=7cm]
			  	\path[->, thick] (0,0) edge (\legendHspace,0);
			  	\node[anchor=west] at (\legendHspace,0) {Jobs precedence ($\graphPrecArcs$)};
			  	\node[anchor=west] at (\legendHspace,-0.5) {+ scheduling precedence choices};
			  	\path[->, thick, dashed] (0,\legendVspace) edge (\legendHspace,\legendVspace);
			  	\node[anchor=west] at (\legendHspace,\legendVspace) {Backward jobs precedence};
			  	\node[anchor=west] at (\legendHspace,\legendVspace-.5) {+ backward scheduling prec. choices};
			  	
			  	\path[<->, thick, red] (0,2*\legendVspace) edge (\legendHspace,2*\legendVspace);
			  	\path[<->, thick, RoyalBlue] (0,2*\legendVspace+.22) edge (\legendHspace,2*\legendVspace+.22);
			  	\path[<->, thick, ForestGreen] (0,2*\legendVspace-.22) edge (\legendHspace,2*\legendVspace-.22);
			  	\draw [decorate,
			  	decoration = {brace},thick] (\legendHspace+.1,2*\legendVspace+.3) --  (\legendHspace+.1,2*\legendVspace-.3);
			  	\node[anchor=west] at (\legendHspace+.2,2*\legendVspace) {Machines conflicts};
			  	
			  \end{scope}
  \end{tikzpicture}
  \centering
  \caption{Rewired graph example with precedences, backward precedences and conflicts as cliques. Each type of arc on the right has its own encoding. Operations $\opIJ{1}{1}$, $\opIJ{2}{1}$,  $\opIJ{2}{2}$ and  $\opIJ{3}{1}$ have here been scheduled in this order.}
  \label{fig:enriched}
\end{figure}

\paragraph{Edge attributes.}
They are used to give explicitly a different type to edges, allowing the network to learn to pass different messages for reverse precedence, precedence and conflicts links. This helps the GNN to effectively handle interactions between tasks of different jobs that share machines.

\paragraph{Node attributes.}
We define for node $n_{ij}$ associated with task $O_{ij}$ the following attributes:
a boolean $A_{ij}$ indicating if the  corresponding task  has already been scheduled (affected);
a boolean $\mi{Sel}_{ij}$ indicating if the node is selectable and the
machine identifier $M_{ij}$.
We also give parameters of probabilistic distribution of tasks durations, and corresponding task completion time distribution parameters.
Task completion times are initialized as if there were no conflicts, i.e. using only durations of task and previous one from same job. There are updated once tasks are affected, considering conflicts with previously affected tasks.

\paragraph{Graph Pooling.}

For the GNN to give a global summary of nodes, there are two options:
either a global isotropic operator on nodes like mean, maximum or sum (or any combination); or a special node that is connected to every task nodes. 
The latter case is equivalent to learning a custom pooling operator.

\paragraph{Output of the GNN.}

The message-passing GNN yields a value for every node, and a global value for the graph (either from the special node or from the chosen isotropic operator). As nodes represent tasks, these values can be directly used as values for later action selection. In our implementation, we also concatenate the global value to every node  value.

\paragraph{Ability to Deal with Different Problem Sizes.}
The GNN outputs a logit for each node, and there is a one-to-one mapping between nodes and actions, whatever the number of nodes/actions. Internally, the  message passing scheme collects messages from all neighbors, making the whole pipeline agnostic to the number of nodes. Learning best actions boils down to node regression, with target values being given by the reinforcement learning loop. This still needs some careful implementation with respect to data structures and batching, but the  direct mapping from nodes to actions allows to deal with different problem sizes.

\subsection{Handling Uncertainty}

 On the observation/agent side, we have durations defined with $\proctimeProbaIJ{i}{j}$. From these durations distributions, we can compute approximate distributions of tasks completion time simply by propagating completion time parameters recursively upon the precedence graph, whenever precedences are added. 
The true real duration of the full schedule is computed only once the complete schedule is known based on all $\proctimeProbaIJ{i}{j}$. It is then passed as a cost signal. As the RL algorithm naturally handles uncertainty of MDPs, it learns to evaluate partial schedule quality based on expectation of costs, which is exactly our objective.

\subsection{Implementation details}

\paragraph{Connecting to PPO.}
In most generic PPO implementation, the actor (policy) consists of a feature extractor whose structure depends on the data type of the observation, followed by a MLP with a output dimension matching the number of actions. Same holds for the critic (value estimator), with the difference that the output dimension is 1. Some layers can be shared (the feature extractor and first layers of the MLPs). In our case, we do not want to use such generic structure because we have a one-to-one matching from the number of nodes of the observation to the actions. We thus always keep the number of nodes as a dimension of the data tensors.

\paragraph{Graph embedder}
The graph embedder builds the rewired graph by adding edges as stated in section \ref{sec:graph-embedding}. It embeds node attributes using a  learnable MLP, and edge attributes (here type of edge only) using a learnable embedding. The output dimension of embeddings is an open hyper-parameter \emph{hidden\_dim}, we found a size of 64 being good in our experiments.

\paragraph{Message-passing GNN}
As a message passing GNN, we use \emph{EGATConv} from the DGL library \cite{dgl}, which enriches GATv2 graph convolutions \cite{gatv2} with edges attributes. We used 4 attention heads, leading to output of size $4  \times $ \emph{hidden\_dim}. This dimension is reduced to \emph{hidden\_dim} using  learnable MLPs, before being passed to next layer (in the spirit of feed-forward networks used in  transformers). This output of a layer can be summed with the input of the layer, using residual connections. For most of our experiments, we used 10 such layers. 

\paragraph{Action selection} 
Action selection aims at giving action probabilities given values (logits) output from the GNN. We can either use logits output by the last layer, or use a concatenation of logits output from every layer. We furthermore concatenate the global graph logits of every layer, leading to a data size of $((n\_layers + 1) \times \emph{hidden\_dim}) \times 2$ per node. This dimension is reduced to 1 using a  learnable linear combination (minimal case of a MLP; we did not find using a full MLP to be useful). Finally, a distribution is built upon these logits by normalizing them, taking into account  action masks at this point. As node numbers correspond to action numbers, we directly have action identifier when drawing a value from the distribution.

\paragraph{Normalization} Along all neural network components, we did not find any kind of normalization to be useful. On the opposite hand, durations are normalized in the [0,1] range. 

\section{Experiments}
\label{sec:experiments}

\subsection{Uncertainty Modeling}

The framework presented in this paper could accommodate to any kind of duration probability distribution. The main parameters of this distribution belong to the node features and must therefore been described formally.

In order to deal with duration uncertainties, for each operation $O_{ij}$, we use a  triangular distribution with 3 parameters $min^p_{ij}$, $max^p_{ij}$, $mode^p_{ij}$, as this is often used in the context of manufacturing processes.

We also have in the simulator the real processing time of a task, denoted $real^p_{ij}$, which is observed by the agent in the final state. With such definition, tasks completion times can respectively be represented with their min, max and mode times as follows: $min^C_{ij} = S_{ij} + min^p_{ij}$, $max^C_{ij} = S_{ij} + max^p_{ij}$, $mode^C_{ij} = S_{ij} + mode^p_{ij}$.  Task completion time parameters are updated when adding precedences to the graph (min, max, mode start times are computed based on precedence relations).  We give both duration distribution parameters and task completion times distribution parameters as node attributes.  The real task completion times $real^C_{ij} = S_{ij} + real^p_{ij}$ can be computed only during the simulation of a complete schedule, giving the real  makespan $M_{real} = max_{ij}(real^C_{ij}$), used as a cost given to the learning agent.

\subsection{Benchmarks and Baselines}

\subsubsection{Benchmarks}
Our approach has been tested on instances generated using Taillard rules \cite{taillard1993benchmarks}: durations are uniformly drawn in  [1,99], and machine affectation is randomly chosen.
For \textit{stochastic} instances, this duration corresponds to $mode^p_{ij}$. Minimum and maximum value are uniformly drawn  in $[0.95 \times mode^p_{ij}, 1.1 \times  mode^p_{ij}]$, meaning that tasks can take at most 5\% less time and at most 10\% more time than mode value.

\subsubsection{Baselines}
In order to evaluate the performance of Wheatley, we compare several approaches:
\begin{itemize}
	\item we first train Wheatley on instances of various sizes. More precisely, \wheatNM{n}{m} denotes our approach tested on instances of size $\ins{n}{m}$, with $(n,m) \in \{(6,6),(10,10),(15,15)\}$;
	\item for deterministic instances, we can compare with L2D using values reported in the associated paper (\cite{L2D});
	\item we test several popular Priority Dispatch Rules (\cite{sels2012comparison}), namely Most Operations Remaining (MOPNR), Shortest Processing Time (SPT), Most Work Remaining (MWKR) and Minimum Ratio of Flow Due Date to Most Work Remaining (FDD/WKR). When computing a schedule, these rules use the mode duration of operations. Next, we retrieve the operations sequence scheduled for each machine and run this sequence with the real operation duration, as done in Schedule Generation Scheme (SGS); 
	\item we use the CP-SAT solver of OR-Tools, denoted \textit{OR-Tools} in deterministic instances test. For stochastic instances, we use it with mode durations and with real instances, respectively denoted \textit{OR-Tools mode} and \textit{OR-Tools real}. As for PDRs, we retrieve the order on each machine and use SGS;
	\item for stochastic instances, we implement the approach proposed in \cite{LocalSolverStocJSSP}, here denoted \textit{CP-stoc}, that consists in finding the schedule that minimize the average makespan over a given number of sampled instances. We found using 50 samples was a very good compromise between solution quality and computation time (100 gives not much improvement and needs too much time).  It is implemented with CP Optimizer 22.10 through docplex (\cite{laborie2018ibm}).
\end{itemize}

Classical techniques like \textit{CP-stoc} and \textit{OR-Tools/CP-sat} are anytime algorithms that need to compute solution for every problem, while our approach uses a large offline training time and the resulting agent only takes a small inference time for every problem. We decide to give 3 minutes to classical techniques, as they tend to give very quickly very good solutions, including for large problems, but generally need up to hours to find optimal solution as soon as the problem size becomes large. On the opposite hand, Wheatley takes from 1 hour to a few days of training (depending on the problem size), but has a fixed inference time that can become very small when correctly optimized (linear in the number of tasks).
The number of iterations to reach the best model is given as table \ref{tab:best_model_n}.
\begin{table}[t!]
  \centering
  \small
  \setlength{\tabcolsep}{5pt}
  \begin{tabular}{c|ccc}
    & \wheatNM{6}{6} & \wheatNM{10}{10} & \wheatNM{15}{15} \\
    \hline
    deterministic & 962 & 542 & 519\\
    stochastic & 712 & 714& 434\\
    \bottomrule
    \end{tabular}
  \caption{Epoch number for best model}
  \label{tab:best_model_n}
  \end{table}

\subsection{Results}

\begin{table}[t!]
	\centering
	\small
	\setlength{\tabcolsep}{3pt}
	\begin{tabular}{@{}cccc@{\qquad}ccc@{}}
		& \multicolumn{3}{c}{Deterministic} &  \multicolumn{3}{c}{Stochastic}  \\
		\cmidrule(r){2-4} \cmidrule(r){5-7}
		Evaluation     & \wheatNM{6}{6}      & \wheatNM{10}{10}    & \wheatNM{15}{15}    &  \wheatNM{6}{6}     & \wheatNM{10}{10}    & \wheatNM{15}{15}  \\ \midrule
		$6\times6$ & $\mf{508} $ & $521 $ & $521 $ & $\mf{700}$ & $714 $ & $715$ \\
		$\ins{10}{10}$ & $927$    & $\mf{890}$    & $915$    & $1269$  & $\mf{1217}$ & $1232$  \\
		$15\times15$ & $1557$   & $\mf{1388}$   & $1392$   & $2297$  & $\mf{1889}$ & $\mf{1889}$  \\
		$20\times15$ & $1798$   & $\mf{1583}$   & $1622$   & $2585$ & $\mf{2181}$ & $2188$  \\
		$20\times20$ & $2314$   & $1959$   & $\mf{1888}$   & $3632$ & $2643$ & $\mf{2608}$  \\
		\bottomrule
	\end{tabular}
	\caption{Comparison of Wheatley wrt training instance sizes.}
	\label{tab:comparison-wheatley}
\end{table}

\subsubsection{Wheatley baselines}
We first compare the three Wheatley baselines together: we have tested them on small instances, both deterministic and stochastic.

Table~\ref{tab:comparison-wheatley} presents results obtained for deterministic and stochastic instances on Taillard problems of several sizes. For each size $\ins{n}{m}$, we have generated 100 instances and, for the stochastic evaluation, we have then sampled one duration scenario for each instance. We then compute the average makespan for each set of instances sizes. Results show that \wheatNM{10}{10} is a good compromise, both for deterministic and stochastic problems. Therefore, in the following, we only present results associated with this approach.

\subsubsection{Deterministic JSSP}

We compare \wheatNM{10}{10} with baselines presented previously for the deterministic case. In Table~\ref{tab:det-taillard-same-size}, we present the average makespan and the average gap\footnote{Gap for an approach $a$ is equal to $100\cdot\frac{\mi{makespan}(a) - \mi{makespan}^\mi{best}}{\mi{makespan}^\mi{best}}$.} obtained for all instances of each category size. Note that we do not present results obtained for each PDR but only the best result one.

\begin{table}[t!]
\centering
\small
\setlength{\tabcolsep}{3pt}
\begin{tabular}{@{}ccccc@{}}
	Evaluation    & \wheatNM{10}{10} & L2D           & Best PDR      & OR-Tools             \\ \midrule
	$6\times6$    & $521~(7.4)$    & $571~(17.7)$  & $545~(12.4)$  & $\mf{485}~(\mf{0})$  \\ 
	$\ins{10}{10}$  & $890~(9.6)$    & $993~(22.3)$  & $948~(16.8)$  & $\mf{812}~(\mf{0})$  \\ 
	$15\times15$  & $1389~(17.2)$  & $1501~(26.7)$ & $1419~(19.8)$ & $\mf{1185}~(\mf{0})$ \\ 
	$20\times15$  & $1583~(16.9)$  & -             & $1642~(21.3)$ & $\mf{1354}~(\mf{0})$ \\ 
	$20\times20$  & $1959~(24.9)$  & $2026~(29.2)$ & $1870~(19.3)$ & $\mf{1568}~(\mf{0})$ \\ 
	$30\times10$  & $1829~(5.5)$   & -             & $1878~(8.9)$  & $\mf{1725}~(\mf{0})$ \\ 
	$30\times15$  & $2043~(14.5)$  & -             & $2092~(17.3)$ & $\mf{1784}~(\mf{0})$ \\ 
	$30\times20$  & $2377~(22.0)$  & -             & $2331~(19.7)$ & $\mf{1948}~(\mf{0})$ \\ 
	$50\times15$  & $3060~(8.3)$   & -             & $3079~(9.0)$  & $\mf{2825}~(\mf{0})$ \\ 
	$50\times20$  & $3322~(14.9)$  & -             & $3295~(14.0)$ & $\mf{2891}~(\mf{0})$ \\ 
	$60\times10$  & $3357~(1.7)$   & -             & $3376~(2.3)$  & $\mf{3301}~(\mf{0})$ \\ 
	$100\times20$ & $5886~(6.9)$   & -             & $5786~(5.1)$  & $\mf{5507}~(\mf{0})$ \\ \bottomrule
	
\end{tabular}
\caption{Results on \textit{deterministic} Taillard instances}
\label{tab:det-taillard-same-size}
\end{table}

Results show that OR-Tools  outperforms the other approaches for these sizes, but \wheatNM{10}{10} manages to get close results, even for large instances. More precisely, in comparison with L2D, which is also an approach based on DRL and GNN that was developed for solving JSSPs (\cite{L2D}), \wheatNM{10}{10} returns better schedules in average. \wheatNM{10}{10} competes with the best PDR, which is mostly MOPNR in the case of instances larger than $\ins{20}{20}$. These results show that  Wheatley is able to learn task selection strategies that generalize to much larger problems.

\subsubsection{Stochastic JSSP}

\begin{table}[t!]
  \centering
  \small
  \setlength{\tabcolsep}{5pt}
\begin{tabular}{@{}ccccccc@{}}
  &  &  &  & & \multicolumn{2}{c}{OR-Tools}  \\
   \cmidrule{6-7}
Evaluation & \wheatNM{10}{10} & \wheatdNM{10}{10}             & MOPNR         & CP-stoc       & mode       & real                 \\ \midrule
$6\times6$      & $714~(16.3)$ & $817~(33.1)$&$699~(13.8)$ & $\mf{669}~(9.0)$ & $728~(18.6)$ & $\mi{614}~(\mi{0})$ \\
$\ins{10}{10}$      & $1217~(21.5)$ &$1464~(46.1)$ & $1252~(25.0)$ & $\mf{1177}~\mf{(17.5)}$ & $1262~(25.9)$ & $\mi{1002}~(\mi{0})$ \\
$15\times15$      & $1889~(29.3)$ & $2406~(64.7)$& $1988~(36.1)$ & $\mf{1872}~\mf{(28.1)}$ & $1925~(31.8)$ & $\mi{1461}~(\mi{0})$ \\
$20\times15$      & $\mf{2181}~\mf{(30.5)}$ & $2729~(63.3)$&$2314~(38.5)$ & $2222~(33.0)$ & $2244~(34.3)$ & $\mi{1571}~(\mi{0})$ \\
$20\times20$      & $2643~(36.4)$ & $3511~(81.2)$& $2708~(40.0)$ & $\mf{2631}~\mf{(35.8)}$ & $2619~(35.1)$ & $\mi{1938}~(\mi{0})$ \\

$30\times10$      & $\mf{2425~(14.1)}$ & $3511~(65.2)$& $2532~(19.1)$ & $2476~(16.5)$ & $2598~(22.2)$ & $\mi{2126}~(\mi{0})$ \\
$30\times15$      & $\mf{2792~(26.7)}$ & $3251~(47.5)$& $2964~(34.5)$ & $2892~(31.2)$ & $2943~(33.5)$ & $\mi{2204}~(\mi{0})$ \\
$30\times20$      & $\mf{3305~(36.9)}$ & $4186~(73.3)$& $3390~(40.4)$ & $3355~(39.0)$ & $3299~(36.6)$ & $\mi{2415}~(\mi{0})$ \\
$50\times15$      & $\mf{4043~(16.5)}$ & $4413~(27.1)$& $4262~(22.8)$ & $4239~(22.1)$ & $4435~(27.7)$ & $\mi{3472}~(\mi{0})$ \\
$50\times20$      & $\mf{4520~(26.8)}$ & $5351~(50.1)$& $4679~(31.2)$ & $4682~(31.3)$ & $4758~(33.4)$ & $\mi{3566}~(\mi{0})$ \\
$60\times10$      & $\mf{4315~(6.3)}$  & $4475~(10.2)$& $4451~(9.6)$  & $4442~(9.4)$  & $4579~(12.8)$ & $\mi{4061}~(\mi{0})$ \\
$100\times20$     & $\mf{7591~(11.8)}$ & $8377~(23.3)$ & $7956~(17.1)$ & $8203~(20.8)$ & $8188~(20.5)$ & $\mi{6793}~(\mi{0})$ \\ \bottomrule
\end{tabular}
\caption{Results on \textit{stochastic} Taillard instances}
\label{tab:stoch-taillard-big-size}
\end{table}

Table~\ref{tab:stoch-taillard-big-size} shows results obtained for stochastic problems. Note that the solver \textit{OR-Tools real} is perfect, in the sense that it works with real operations duration values, which is unknown for other approaches at the scheduling time. Therefore, the makespan value computed by \textit{OR-Tools real} is much lower than that of other approaches. Results show that the closest to \textit{OR-Tools real} is \textit{CP-stoc} for small problem sizes. In fact, despite the 50 scenarios it works with, it manages to find a good average makespan. However, when the instances size increases, \wheatNM{10}{10} clearly outperforms other approaches. We also present the results for the deterministic of version of Wheatley run on modes as \wheatdNM{10}{10}.
This shows that Wheatley is able to successfully generalize on larger problems.

\begin{figure}[ht!]
	\begin{subfigure}[t]{.50\textwidth}
		\includegraphics[width=6.5cm]{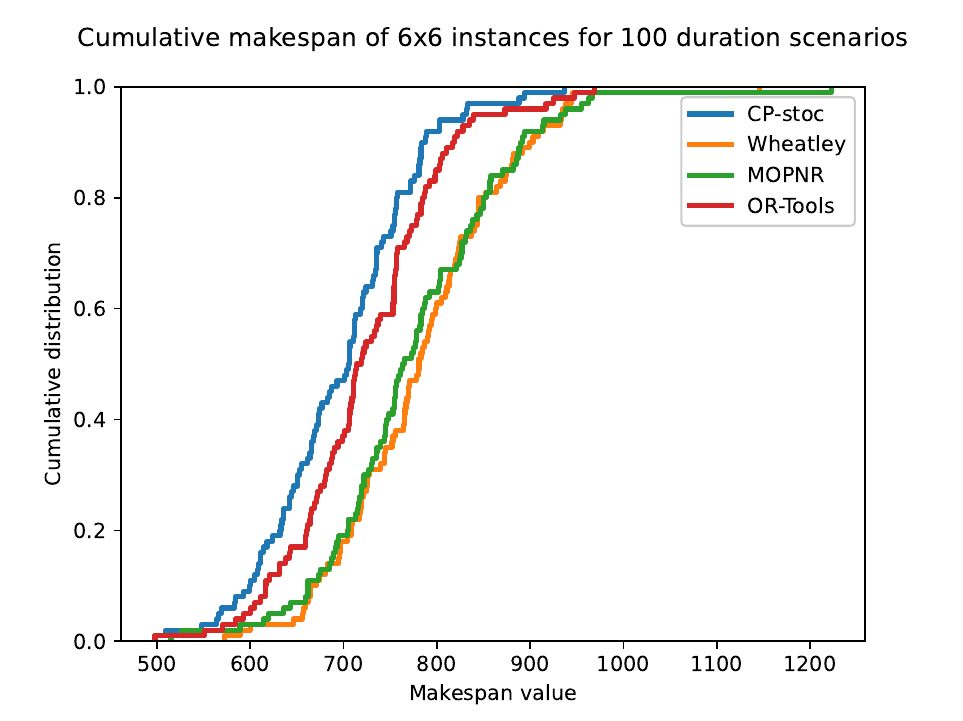}
		\caption{$6 \times 6$ instance}
		\label{fig:cumul_6}
	\end{subfigure}
	\hfill
	\begin{subfigure}[t]{.50\textwidth}
		\includegraphics[width=6.5cm]{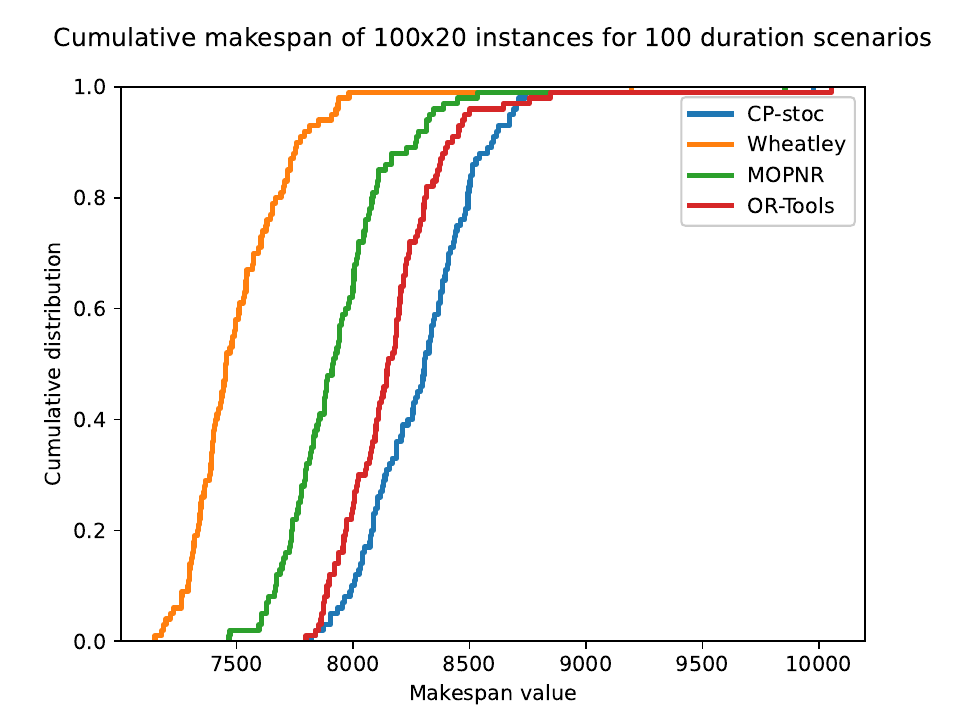}
		\caption{$100 \times 20$ instance}
		\label{fig:cumul_100}
	\end{subfigure}
	\caption{Cumulative makespan of \wheatNM{10}{10} and \textit{CP-stoc} for 100 duration scenarios.}
	\label{fig:cumul}
\end{figure}

In order to further compare the approaches in terms of scatter of the results, we have sampled 100 duration scenarios for one problem of size $6 \times 6$ and 100 scenarios for one problem of size $100 \times 20$. Cumulative makespan are presented on Figure~\ref{fig:cumul}. It shows that results presented on Table~\ref{tab:stoch-taillard-big-size} are representative of several scenarios. In fact, for the $\ins{6}{6}$ problem, \textit{CP-stoc} returns the lowest makespans, then \textit{OR-Tools}, and \textit{MOPNR} and \wheatNM{10}{10} equivalently (Figure~\ref{fig:cumul_6}). That order is completely reversed in the case of the $\ins{100}{20}$ problem, in which \wheatNM{10}{10} returns the best results (Figure~\ref{fig:cumul_100}).

\section{Conclusion and Future Works}
\label{sec:conclusion}
This paper presents Wheatley, a novel approach for solving JSSPs with uncertain operations duration. It combines Graph Neural Networks and Deep Reinforcement Learning techniques in order to learn a policy that iteratively selects the next operation to execute on each machine. The policy is updated during the training phase through PPO. Results show that Wheatley is competitive in the case of deterministic JSSPs and  outperforms other approaches for stochastic JSSPs. Moreover, Wheatley is able to generalize to larger instances.

This work could be extended in several directions. First, it would be possible to extend the experiments with other JSSPs data and particularly instances coming from the industry. It would also be interesting to study the effect of pretraining the policy before running PPO. Finally, we are convinced that the GNN and DRL could be applied to other scheduling problems, such as the Resource-Constrained Project Scheduling Problem, in which handling uncertainty is essential in an industrial context.

\newpage
\bibliographystyle{plain}
\bibliography{wheatley_jssp}

\end{document}